\documentclass{article}

\usepackage{microtype}
\usepackage{graphicx}
\usepackage{subcaption}
\usepackage{booktabs} % for professional tables

\usepackage{hyperref}

\usepackage[accepted]{icml2021}

\usepackage{amsmath,amssymb}
\usepackage{multirow}

\icmltitlerunning{Adversarial Defense via Image Denoising with Chaotic Encryption}

\begin{document}

\twocolumn[
\icmltitle{Adversarial Defense via Image Denoising with Chaotic Encryption}




\begin{icmlauthorlist}
\icmlauthor{Shi Hu}{uva}
\icmlauthor{Eric Nalisnick}{uva}
\icmlauthor{Max Welling}{uva}
\end{icmlauthorlist}

\icmlaffiliation{uva}{University of Amsterdam}

\icmlcorrespondingauthor{Shi Hu}{s.hu@uva.nl}

\icmlkeywords{Adversarial Robustness, Deep Learning, Chaotic Encryption}

\vskip 0.3in
]



\printAffiliationsAndNotice{}  % leave blank if no need to mention equal contribution

\begin{abstract}

In the literature on adversarial examples, white box and black box attacks have received the most attention.  The adversary is assumed to have either full (white) or no (black) access to the defender's model. In this work, we focus on the equally practical gray box setting, assuming an attacker has partial information.  We propose a novel defense that assumes everything but a private key will be made available to the attacker. Our framework uses an image denoising procedure coupled with encryption via a discretized Baker map. Extensive testing against adversarial images (e.g.~FGSM, PGD) crafted using various gradients shows that our defense achieves significantly better results on CIFAR-10 and CIFAR-100 than the state-of-the-art gray box defenses in both natural and adversarial accuracy.

\end{abstract}

\section{Introduction}
For our AI systems to be considered safe, we must ensure that they can withstand attacks from adversaries.  Unfortunately, most machine learning models are vulnerable to these attacks \cite{szegedy}.  Consider the task of autonomous driving.  These systems can not only be attacked virtually, by directly manipulating inputs such as pixels, but they can also be attacked by altering the physical world.  For instance, \citet{eykholt} show that stop signs can be erroneously classified as speed limit signs simply by placing a small patch on the sign.

Protecting our systems from such attacks is made especially challenging when there is a need for real-time decision making.  Again consider a vision system for an autonomous vehicle.  Since objects on the road must be detected as quickly as possible, there is no time to communicate with a cloud service, and the classifier must be located in on-device memory.  Hence, an adversary will have direct access to the on-device model, allowing them to craft adversarial examples quite easily \cite{kurakin}.  To address these cases, we must develop safety methodologies that assume the model is visible to the attacker.

\begin{figure*}[h]
    \centering
    \begin{subfigure}[b]{\columnwidth}
        \centering
        \includegraphics[width=0.7\textwidth]{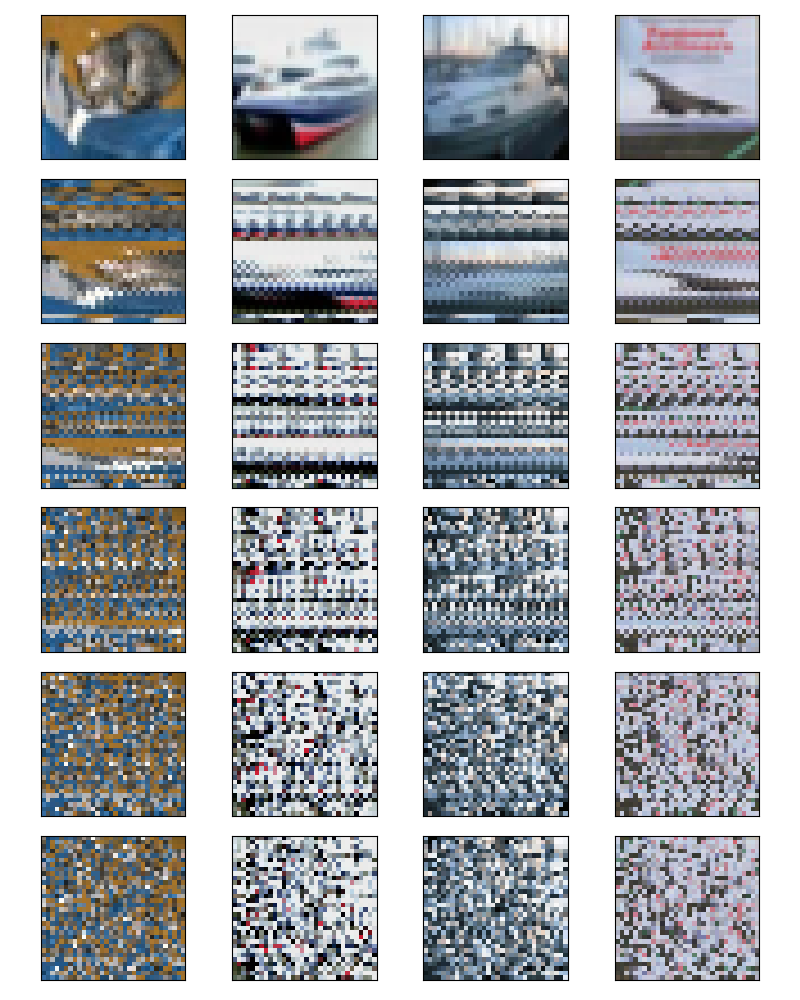}
        \caption{$\texttt{k}_1=(2, 16, 8, 4, 1, 1)$.}
    \end{subfigure}
    \hfill
    \begin{subfigure}[b]{\columnwidth}
        \centering    
        \includegraphics[width=0.7\textwidth]{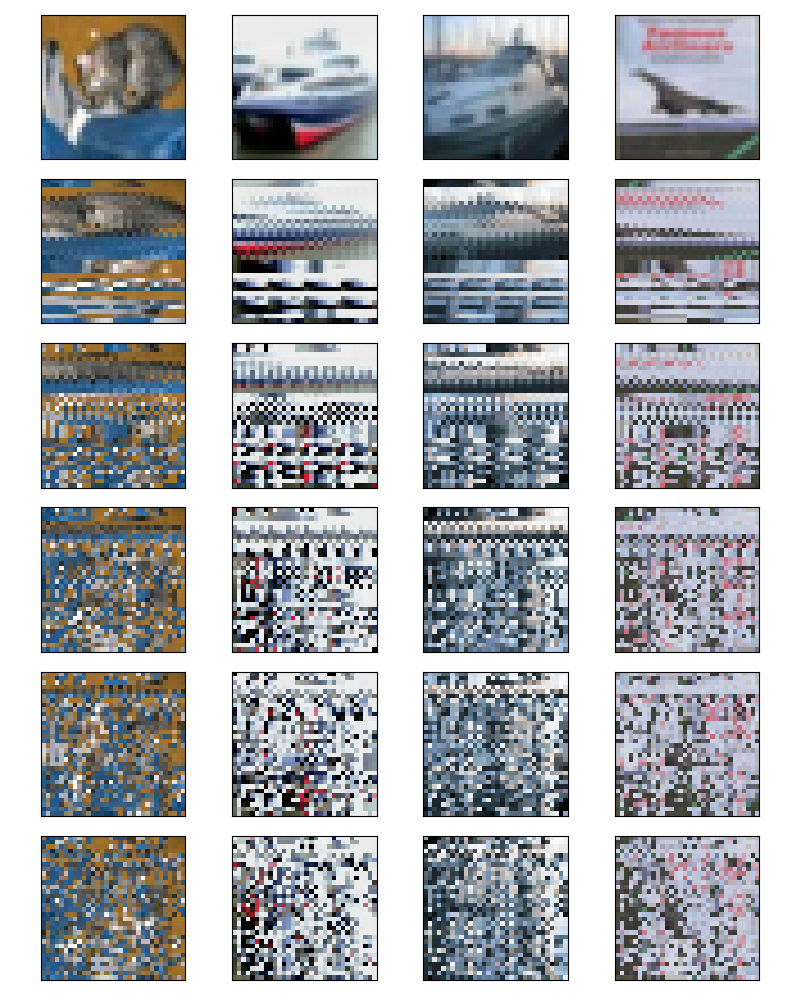}
        \caption{$\texttt{k}_2=(2, 1, 4, 4, 1, 1, 1, 16, 1, 1)$.}
    \end{subfigure}        
    \caption{The Baker map encrypted CIFAR-10 images under keys $\texttt{k}_1$ and $\texttt{k}_2$, with 1 to 5 iterations.}
    \label{fig:baker_iters}
\end{figure*}

\begin{figure*}[h]
    \centering
    \includegraphics[width=0.8\textwidth]{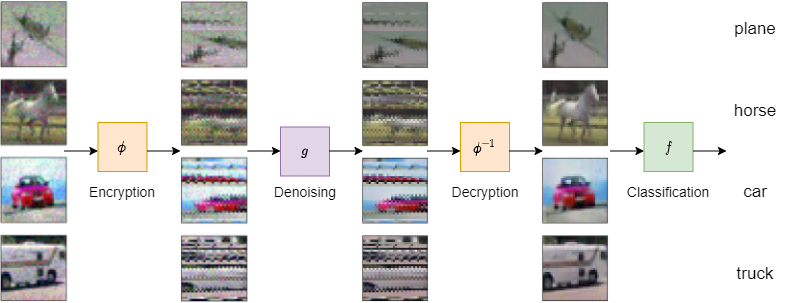}
    \caption{A diagram of our defense via image denoising with the Baker map encryption. The inputs are adversarial CIFAR-10 images.}
    \label{fig:workflow}
\end{figure*}

We propose a novel \emph{gray box} adversarial defense that assumes everything but a private key will be made available to the attacker.  We accomplish this by using an image denoising procedure coupled with encryption via a discretized Baker map \cite{fridrich}.  The encryption key is what we must keep secret from the attacker.  For a sketch of our approach, Figure \ref{fig:baker_iters} demonstrates the Baker map encryption, and Figure \ref{fig:workflow} shows a diagram of our proposed method.  We evaluate our defense against two gradient-based adversarial attacks: the \textit{Fast Gradient Sign Method} \cite{fgsm} and \textit{Projected Gradient Descent} \cite{pgd}.  We achieve significantly better results than state-of-the-art gray box defenses in both natural and adversarial accuracy.

\section{Setting of Interest: Gray Box Attacks}
\paragraph{Notation} Throughout the paper, let $x$ denote an image, $y(x)$ its true label, $f$ a classifier, $g$ an image denoiser, $\texttt{k}$ an encryption key, $\phi$ and $\phi^{-1}$ an image encryption and its decryption algorithm, and $\ell\big(f(x), y(x)\big)$ a loss function (e.g., the cross-entropy loss). We abbreviate $\ell\big(f(x), y(x)\big)$ as $\ell(f(x))$, \textit{Fast Gradient Sign Method} as FGSM, and \textit{Projected Gradient Descent} as PGD. PGD-$q$ means PGD uses $q$ optimization steps.

We begin by defining the classes of adversarial attacks: \textit{white}, \textit{gray}, and \textit{black} box.  The distinguishing characteristic is in how much information is assumed available to the attacker.

\textbf{White Box} The white box setting assumes that the attacker has access to the same information as the defender, including the model parameters, training data, and the defense mechanism (e.g., adversarial training \cite{fgsm}).  This setting places the most burden on the efficacy of the defense mechanism, with the upside being that there is no need to keep information private upon deployment.

\textbf{Gray Box}  The gray box setting assumes that only \emph{partial} information about the model is available to the attacker.  The inaccessible data can include the trainable parameters and an encryption key \cite{taran}. In this paper, we assume the key is always private (if it is used), and further subdivide the gray box attacks into \textit{dark}, \textit{medium} and \textit{light} types, where the trainable parameters are completely inaccessible, partially inaccessible, and completely accessible.

\textbf{Black Box}  The black box setting assumes that all information about the model is \emph{unknown} to the attacker.  The attacker can proceed only by training a surrogate model using the target model's outputs as labels \cite{papernot}.  This setting places the most burden on the system's security since all information must be protected.  If the attacker is able to bypass this security and access the model, then it will be easy to degrade the system's performance since no other protections are in place.

\paragraph{Motivation} In this paper, we focus on the \textbf{gray box setting}.  The white box and black box settings have attracted the most attention in the literature. In comparison, gray box attacks are under-explored yet are still germane to many modern technologies.  Returning to the self-driving car example, the car manufacturer is free to publish (via software update) their architecture and/or defense mechanism without a concern for security, as long as the designated private information is kept as such (e.g. encryption key).  In our approach, the private information will be an encryption key, which should be easier to keep private than the parameters of a large neural network.

%In machine learning, model training usually involves randomness controlled by a pseudorandom number generator, and different random seeds usually lead to non-identical sets of model parameters. The gray box defender can then exploit this property to protect their model by keeping their seed private, such that the attacker cannot access their model parameters, and thus cannot use the exact gradient to craft the adversarial images. For example, the defender can use the seed to initialize their classifier's weights during training \cite{pgd}\footnote{We note that \cite{pgd} refers to this setting as the ``black box'' defense.}; or generate a private key for some public encryption algorithm (the key generator is also public), then uses it to encrypt the input images in training and testing \cite{taran}. These gray box defenses achieve much higher adversarial accuracy compared to the white box counterpart, where the exact gradient is accessible to the attacker. However, their performances are still far from satisfactory for real-world applications; in addition, they impose a considerable cost for natural accuracy. 

Our proposed method is inspired by the limitations of previously proposed denoising- and encryption-based gray box approaches.  For an example of the former, \citet{liao} train a classifier ($f$) using natural images, then use the U-net \cite{unet} as a denoising model ($g$) to protect against the adversarial images crafted using the gradient $\nabla_x \ell(f(x))$. This defense mechanism achieves good natural and FGSM accuracy \cite{liao}. Unfortunately, \citet{liao_attack} show that it is vulnerable to PGD attacks crafted using the end-to-end gradient $\nabla_x \ell\big(f(g(x))\big)$ (we note \cite{liao} was proposed as a white box defense, but their vulnerability to PGD attacks remains in the gray box setting).  For an example of the latter, \citet{taran} encrypt the input images during training and testing.  However, their method's performance is still far from satisfactory in natural accuracy and real-world applications.

\section{Method}

%We first encrypt the input images via a public encryption algorithm $\phi_{\texttt{k}}$ with the private key $\texttt{k}$, then feed the encrypted images to the U-net $g_{\texttt{k}}$ for denoising. We then feed the decrypted outputs $\hat{x}_{\texttt{k}} = \phi^{-1}_{\texttt{k}}(g_{\texttt{k}}(\phi_{\texttt{k}}(x)))$ to the classifier.

%The U-net is a convolutional neural network (CNN) \cite{cnn}, which processes the images by local patches. If the pixels in an encrypted image patch are totally random and not correlated with each other, it is challenging for the CNN to denoise these pixels. In addition, if the encrypted images used by the attacker and defender are not sufficiently different, the attacker can still use the end-to-end gradient to attack the defender. Therefore, given any key, the output of the encryption needs to preserve some local structure of the input; at the same time, different keys need to lead to sufficiently different outputs.

We propose an encrypt-denoise-decrypt approach.  In the first step, we encrypt the images via a public encryption algorithm $\phi_{\texttt{k}}$ with private key $\texttt{k}$.  In the second step, the encrypted images are passed into a denoising model $g_{\texttt{k}}$.  We use the U-net for $g_{\texttt{k}}$, following \cite{liao}.  The third step is to decrypt the denoised image: $\hat{x}_{\texttt{k}} = \phi^{-1}_{\texttt{k}}(g_{\texttt{k}}(\phi_{\texttt{k}}(x)))$ where $\hat{x}_{\texttt{k}}$ denotes the final decypted and denoised image. Lastly, we feed $\hat{x}_{\texttt{k}}$ to the classifier for prediction. Below we discuss the encryption and denoising procedures in more detail.

\subsection{Chaotic Encryption}
We need the encryption algorithm to meet two criterion.  The first is that distinct keys must result in sufficiently different outputs.  If many keys were to result in the same encrypted images, then the key need not be compromised for an attack to succeed.  Rather, the attacker needs only to guess one of the equivalent keys.  The second requirement is a result of using a CNN-based denoiser.  Because CNNs operate by way of local correlations, the encryption algorithm must preserve the input's local structure to some degree.

For our choice of the encryption algorithm $\phi_{\texttt{k}}$, we use the discretized Baker map proposed by \cite{fridrich}, version A. We consider only square images in this work, but the encryption can be extended to rectangular images \cite{fridrich}. Further, if the image has multiple channels, we apply the same encryption to each channel. The key for the encryption is a small set of integers $\texttt{k}=(n_1, n_2, \dots, n_m)$, where each $n_i$ is divisible by the image dimension $N$, and their sum $\sum_{i=1}^m n_i = N$. Let us denote $N_i = \sum_{j=1}^i n_i$, where $N_0 = 0$. Then, for the pixel at position $(w_t, h_t)$ at iteration $t$, where $N_{i-1} \le w_t < N_{i}$ and $0 \le h_t < N$, we use the following equations to move it to the new position $(w_{t+1}, h_{t+1})$:

\begin{align}
    w_{t+1} &= \frac{N}{n_i} (w_t - N_i) + (h_t \mod \frac{N}{n_i}), \\
    h_{t+1} &= \frac{n_i}{N}\big(h_t - (h_t \mod \frac{N}{n_i})\big) + N_i.
\end{align}

The key $(n_1, n_2, \dots, n_m)$ partitions the image into $m$ vertical rectangles, each of shape $N \times n_i$. Further, each rectangle is partitioned into $n_i$ boxes, where each box has shape $\frac{N}{n_i} \times n_i$, and has thus exactly $N$ pixels. Next, we stretch the pixels in each box to a row of shape $1 \times N$, then stack the rows together to produce the permuted image. Two examples of the encryption using keys $(1,1,2)$ and $(1,2,1)$ on a simple $4 \times 4$ matrix are shown in Eqs. \ref{eq:perm112} and \ref{eq:perm121}. 

\begin{align}
    \begin{array}{@{}c|c|cc@{}}
        0 & 1 & 2 & 3 \\
        4 & 5 & 6 & 7 \\ \cline{3-4}
        8 & 9 & 10 & 11 \\
        12 & 13 & 14 & 15
    \end{array}
    &\quad \xrightarrow[]{\phi_{(1,1,2)}} \quad     
    \begin{array}{@{}cccc@{}}
        6 & 2 & 7 & 3 \\ \hline
        14 & 10 & 15 & 11 \\ \hline
        13 & 9 & 5 & 1 \\ \hline
        12 & 8 & 4 & 0 
    \end{array}
    \label{eq:perm112} \\
    \begin{array}{@{}c|cc|c@{}}
        0 & 1 & 2 & 3 \\
        4 & 5 & 6 & 7 \\ \cline{2-3}
        8 & 9 & 10 & 11 \\
        12 & 13 & 14 & 15
    \end{array}       
    &\quad \xrightarrow[]{\phi_{(1,2,1)}} \quad    
    \begin{array}{@{}cccc@{}}
        15 & 11 & 7 & 3 \\ \hline
        5 & 1 & 6 & 2 \\ \hline
        13 & 9 & 14 & 10 \\ \hline
        12 & 8 & 4 & 0
    \end{array}   
    \label{eq:perm121}
\end{align}

The total number of possible keys $K(N)$ depends on both the image dimension $N$ and the number of divisors of $N$. In general, $K(N)$ grows rapidly with $N$. Table \ref{tab:num_keys} shows a few examples of their relation, which suggests that it is highly unlikely that the attacker and defender can use the same key by chance.

\begin{table}[h]
    \centering
    \begin{tabular}{|c|c|} 
    \hline    
    $N$ & $K(N)$ \\ \hline
     16 & 5271 \\
     32 & $4.7 \times 10^7$  \\
     64 & $3.8 \times 10^{15}$ \\
     128 & $10^{31}$ \\
     256 & $10^{63}$ \\
     512 & $10^{126}$ \\
     1024 & $10^{255}$ \\ \hline
    \end{tabular} 
    \caption{Image dimension $N$ and total number of keys $K(N)$.}
    \label{tab:num_keys}
\end{table}

Let us examine the Baker map encrypted images in Figure \ref{fig:baker_iters}. The figure shows that more encryption iterations leads to more random image patterns, where the local spatial relations in the input images are destroyed. However, with few iterations, some spatial relations are preserved and the resulting images have perceptible structures, yet different keys lead to noticeably different structures. These properties make the Baker map encryption ideally suited for our purposes, though it can be replaced by other viable encryption schemes for our defense.

\subsection{Training Procedure}

Our defense trains the classifier $f$ and the U-net denoiser $g_{\texttt{k}}$ in 2 steps. We first train the classifier using only natural images. After training, we fix its weights, and then train the denoiser using the encrypted adversarial inputs $\phi_{\texttt{k}}(x_{\text{adv}})$. The gradient for the adversarial perturbation comes from the trained classifier, which is $\nabla_x \ell(f(x))$. Let us denote $f_{-1}(\cdot)$ as the feature map after the last convolution block of the classifier, our training loss for the denoiser is:

\begin{align}
    L_1 (\hat{x}_{\texttt{k}}, x_{\text{nat}}) + 
    L_1(f_{-1}(\hat{x}_{\texttt{k}}), f_{-1}(x_{\text{nat}})), \label{eq:unet_loss}
\end{align}

where $\hat{x}_{\texttt{k}} = \phi^{-1}_{\texttt{k}}(g_{\texttt{k}}(\phi_{\texttt{k}}(x_{\text{adv}})))$. Namely, we minimize the $L_1$ distances between the denoised and natural images $\hat{x}_{\texttt{k}}$ and $x_{\text{nat}}$, as well as the last convolution feature maps for the two images. The second $L_1$ loss provides guidance to the first one, and the combined loss achieves much better results than the first loss alone.

Alternatively, we could feed the encrypted images directly to the classifier without the denoiser. If the input images have simple structures such as MNIST, then feeding the permuted images to a CNN-based classifier barely reduces the natural accuracy; however, as the complexity of the dataset grows, classifying the permuted images leads to increasingly lower accuracy compared to the unpermuted counterpart \cite{ivan}. Using the U-net denoiser with image encryption and decryption alleviates this problem: though the U-net is still CNN-based, the denoised images are decrypted (i.e., permuted back) to the original pixel order before being fed to the classifier, which improves the accuracy. Further, since the denoiser is trained only to denoise images, the attacker cannot easily deduce the private key even if they can access the denoiser's weights. In comparison, if the denoiser learns to both denoise and decrypt images, then it is possible that the attacker can recover the private key by feeding it with inputs of simple patterns (e.g., an image that contains a single non-zero column), and observing the pixel orders in the outputs.

\subsection{Adversarial Attacks} \label{sec:attacks}

In gray box attacks, the defender's key generator and model architecture are public, but their random seed is assumed to be different from the attacker's. Thus, their generated private keys and initial model weights are different. 

In testing, the defender's model $f\big(\phi^{-1}_{\texttt{k}}(g_{\texttt{k}}(\phi_{\texttt{k}}(x)))\big)$ with the private key $\texttt{k}$ can be attacked by the adversarial images crafted using the 4 gradients listed below. To compute these gradients, the attacker uses the defender's classifier's weights if their are public; otherwise, the attacker trains their classifier initialized using their random seed. 

\begin{enumerate}
    \item The gradient from the classifier: $\nabla_x \ell(f(x))$.
    \item The end-to-end gradient without encryption: attacker trains their denoiser $g$ using unencrypted images, and obtains the gradient $\nabla_x \ell\big(f(g(x))\big)$.
    \item The end-to-end gradient with encryption using the key for training: the attacker trains their denoiser $g_{\texttt{k}'}$ using encrypted images with their key $\texttt{k}' \ne \texttt{k}$, and obtains the gradient $\nabla_x \ell\big(f(\phi^{-1}_{\texttt{k}'}(g_{\texttt{k}'}(\phi_{\texttt{k}'}(x))))\big)$.
    \item  The end-to-end gradient with encryption using random keys. This is similar to gradient 3, except in testing, rather than using the same key for training, the attacker samples a random key $\tilde{\texttt{k}}$ for FGSM, and each step of PGD-$q$ (i.e., there are $q$ keys for PGD-$q$). Thus, if the defender's denoiser's weights are public, then the attacker obtains the gradient $\nabla_x \ell\big(f(\phi^{-1}_{\tilde{\texttt{k}}}(g_{\texttt{k}}(\phi_{\tilde{\texttt{k}}}(x))))\big)$. Otherwise, the attacker trains their denoiser $g_{\texttt{k}'}$ using the encrypted images with their key $\texttt{k}' \ne \texttt{k}$, and obtains the gradient $\nabla_x \ell\big(f(\phi^{-1}_{\tilde{\texttt{k}}}(g_{\texttt{k}'}(\phi_{\tilde{\texttt{k}}}(x))))\big)$.
\end{enumerate}

\section{Experiments}

We use the original ResNet-18 classifier \cite{resnet}, and the CIFAR-10 and CIFAR-100 datasets \cite{cifar} with the original image dimensions $32 \times 32 \times 3$. For natural training (i.e., the classifier is trained using only natural images), we use the same training procedure as \cite{pgd}, which augments the training inputs using cropping and horizontal flipping, and optimizes the model using stochastic gradient descent (SGD) for 200 epochs. The learning rate starts at $10^{-1}$, and decays to $10^{-2}$ and $10^{-3}$ after 100 and 150 epochs. This achieves 95.28\% and 75.55\% test accuracy on natural CIFAR-10 and CIFAR-100 images respectively. 

We use two gradient-based adversarial attacks in  $\ell_{\infty}$ norm, which are FGSM and PGD-$q$. FGSM is an one-step attack, and the adversarial image is a point on the boundary of a small norm ball centered at $x$ in the direction of the sign of the gradient:

\begin{align}
    x' = x + \epsilon \cdot \text{sign} \big(\nabla_x \ell(f(x))\big). \label{eq:fgsm}
\end{align}

In comparison, PGD-$q$ is a multi-step attack, which starts at a random point within the norm ball $S$ centered at $x$, and runs $q$ steps. In each step $t$, it finds a point $x^{t+1}$ within $S$ that maximizes the loss with the step size $\alpha$:

\begin{align}
    x^{t+1} = \text{Proj}_{x + S} \big[x^t + \alpha \cdot \text{sign} \big(\nabla_{x^t} \ell(f(x^t))\big) \big]. \label{eq:pgd}
\end{align}

For both attacks, the radius of the norm ball $\epsilon = \frac{8}{255}$; for PGD-$q$, the step size $\alpha = \frac{2}{255}$. We use the CleverHans \cite{papernot2018cleverhans} implementation in PyTorch \cite{pytorch}\footnote{\url{https://github.com/cleverhans-lab/cleverhans}} to craft the adversarial images, and compare with two adversarial defenses, which are adversarial training and multi-channel sign permutation. The former is one of the few unbroken defenses to this day, and the latter is an existing state-of-the-art gray box defense with an encryption key.

%For adversarial training, the CleverHans version achieves higher natural and adversarial accuracy compared to the version by Madry et al.\footnote{\url{https://github.com/MadryLab/cifar10_challenge}} The difference is that the former reduces the probability of the predicted class, while the latter that of the true class. 

\subsection{Adversarial Training}

In adversarial training \cite{fgsm,pgd}, the defender trains the classifier $f$ using adversarial images. For each natural image $x_{\text{nat}}$ and its label $y(x_{\text{nat}})$, this defense first creates the adversarial image $x_{\text{adv}}$ using FGSM (Eq. \ref{eq:fgsm}) or PGD-$q$ (Eq. \ref{eq:pgd}), then updates the classifier's weights using the input $x_{\text{adv}}$ and label $y(x_{\text{nat}})$. 
In the experiments, we use PGD-7 for the training inputs. We also tried PGD-20 as the inputs, and found the results are very similar to that of PGD-7, but training takes much longer. Lastly, we train the classifier using the same procedure as for natural training.

%For dark gray box attacks, the attacker cannot access the defender's classifier's weights. Therefore, they use the same training data and procedure to train an identical classifier, which they initialize using their random seed. After training, they obtain its gradient to craft the adversarial images to attack the defender's classifier. 

\subsection{Multi-Channel Sign Permutation}

\citet{taran} propose an encryption-based defense mechanism, which first transforms the input image $x$ using the public discrete cosine transform (DCT) $W$, then multiplies it with a secret sign permutation matrix $P$, and finally transforms it back to the image domain using the inverse DCT $W^{-1}$. There are $J=3$ possible DCT sub-bands, which are vertical (V), horizontal (H) and diagonal (D), and $I$ channels per sub-band. For each channel, it performs the encryption as $\phi_{ji}(x) = W_j^{-1} P_{ji} W_j x$, where $1 \le j \le J,\ 1 \le i \le I$, then classifies the encrypted image $\phi_{ji}(x)$ using a convolutional network $f_{ji}$. Lastly, it averages over the outputs of all channels to make the final prediction $\hat{y}(x) = \frac{1}{JI}\sum_{j=1}^J\sum_{i=1}^I f_{ji}(\phi_{ji}(x))$.

%As mentioned in their paper, this defense is applicable only in the dark gray box setting, where the attacker cannot access the defender's model parameters. In this case, the attacker trains an identical ensemble of classifiers with the same encryption scheme, training data and procedure, but uses their own random seed. Then they use the gradient of this model to craft the adversarial images. 

For the experiments, we follow the same test setup as in Table 2 of their paper. Namely, we use all 3 sub-bands V, H and D, and increase the number of channels per sub-band I from 1 to 5. The total number of channels $JI$ is thus 3, 6, 9, 12 or 15. Lastly, we use the same implementation released by the authors\footnote{\url{https://github.com/taranO/defending-adversarial-attacks-by-RD}}, but re-implement the code in PyTorch to be consistent with other methods.

\subsection{Our Defense}

We use an existing implementation for the U-net denoiser by \citet{fastMRI}\footnote{\url{https://github.com/facebookresearch/fastMRI/blob/14562052eb3f37dd1f23f694bddfc3b8d456d571/models/unet/unet_model.py}}. It has 4 convolution blocks, the number of output channels of the first convolution layer is 128, and dropout \cite{dropout} is not used. Further, we randomly choose the FGSM and PGD-7 adversarial images as its input with equal probabilities. We found that alternating the two types of inputs achieves better overall adversarial accuracy compared to using just one of them. However, for adversarial training, using the PGD-7 inputs alone works better. Lastly, we optimize the denoiser using the same procedure as for the classifier, but we do not use input augmentations.

%For dark gray box attacks, the defender's key and model weights are private. Thus, the attacker uses the same training data and procedure to train their model with their random seed, and uses the 4 gradients listed in Section \ref{sec:attacks} to craft the adversarial images. 

\subsection{Results}

\subsubsection{Dark Gray Box Defense} \label{sec:graybox_results}

\begin{table*}[h]
    \centering
    \begin{tabular}{|l|cccc|cccc|} 
    \hline
    \multirow{2}{*}{Defense} & \multicolumn{4}{c}{CIFAR-10}  & \multicolumn{4}{|c|}{CIFAR-100} \\ 
     & Natural & FGSM & PGD-7 & PGD-20 & Natural & FGSM & PGD-7 & PGD-20 \\ \hline
     Natural training & 95.28 & 47.15 & 10.01 & 7.11 & 75.55 & 18.67 & 11.82 & 9.97 \\ \hline
     Adversarial training & 84.74 & 72.17 & 71.26 & 70.16 & 56.47 & 46.86 & 46.87 & 46.12 \\
     Taran et al. ($JI=3$)  & 80.89 & 60.61 & 56.28  & 50.05 & 46.56 & 39.04 & 42.08 & 40.25 \\
     Taran et al. ($JI=6$)  & 83.08 & 55.58 & 48.87 & 43.10 & 50.37 & 40.26 & 43.37 & 40.41 \\
     Taran et al. ($JI=9$)  & 83.34 & 52.34 & 45.60 & 41.09 & 51.69 & 39.77 & 41.62 & 37.97 \\
     Taran et al. ($JI=12$) & 83.89 & 50.68 & 44.12 & 39.68 & 52.59 & 37.99 & 40.18 & 35.74 \\
     Taran et al. ($JI=15$) & 83.94 & 50.03 & 43.35 & 39.05 & 53.06 & 36.67 & 38.18 & 33.81 \\
     Ours (without encryption) & \textbf{93.58} & \textbf{81.15} & 64.78 & 62.37 & \textbf{70.42} & 53.72 & 49.05 & 44.46 \\
     Ours (with encryption) & 92.43 & 78.96 & \textbf{83.32} & \textbf{77.87} & 70.09 & \textbf{55.95} & \textbf{61.32} & \textbf{59.40} \\
     \hline
    \end{tabular} 
    \caption{The dark gray box test accuracy (in \%). For ours, the minimum accuracy against the 4 gradients for each attack is shown.}
    \label{tab:graybox_all}
\end{table*}

For \textit{dark} gray box attacks, the adversary does not have access to the defender's model parameters or key. Thus, they use the same training data and procedure to train an identical model with their initialization seed, then use the gradient of this model to craft the adversarial images\footnote{We note that \citet{pgd} refer to this setting as the ``black box'' attack.}. Table \ref{tab:graybox_all} compares the test accuracy on CIFAR-10 and CIFAR-100. As a reference, we also show the test results for natural training. Unsurprisingly, it achieves the highest accuracy on natural images, but lowest on adversarial images. In comparison, adversarial training markedly improves the adversarial accuracy, but comes with a considerable cost for natural accuracy. For Taran et al., as the total number of channels $JI$ increases from 3 to 15, the natural accuracy slowly increases, but all adversarial accuracies decrease. In addition, none of the configurations achieves better results than adversarial training. Its natural accuracy is comparable with adversarial training, since the encryption method does not significantly alter the appearance of the input images\footnote{This is illustrated in Figure 6 of \cite{taran}.}. Unfortunately, due to the same reason, the trained weights for the defender's ensemble are not sufficiently different from the attacker's, which makes the former vulnerable to the adversarial attacks.

For our method, we show the results without and with encryption. If encryption is not used, the defender feeds the inputs directly to the denoiser. Otherwise, they first choose a private key $\texttt{k}$ for the Baker map, then feed the encrypted images to the denoiser. Since each type of adversarial image (FGSM, PGD-7 or PGD-20) can be crafted using one of the four gradients described in Section \ref{sec:attacks}, we report the minimum accuracy of the four for each attack. This is because if the attacker knows the defender's model, then they will use the most harmful gradient to craft the adversarial images. For these results, the defender's key $\texttt{k} = (2, 16, 8, 4, 1, 1)$, and the attacker's key for gradient 3 or 4 is $\texttt{k}' = (2, 1, 4, 4, 1, 1, 1, 16, 1, 1)$. Both keys are randomly chosen.

\begin{figure}[h]
    \centering
    \includegraphics[width=\columnwidth]{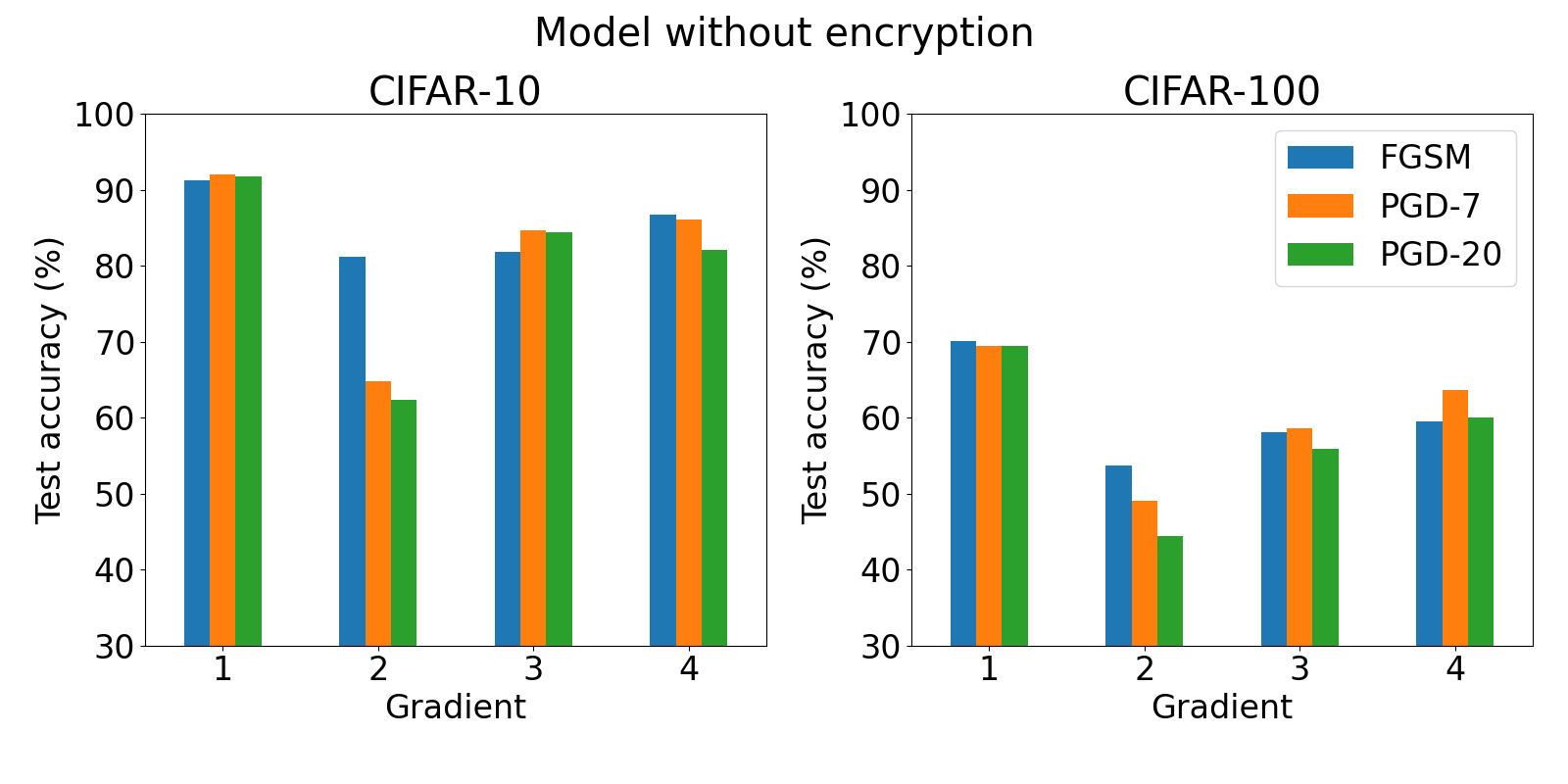}
    \includegraphics[width=\columnwidth]{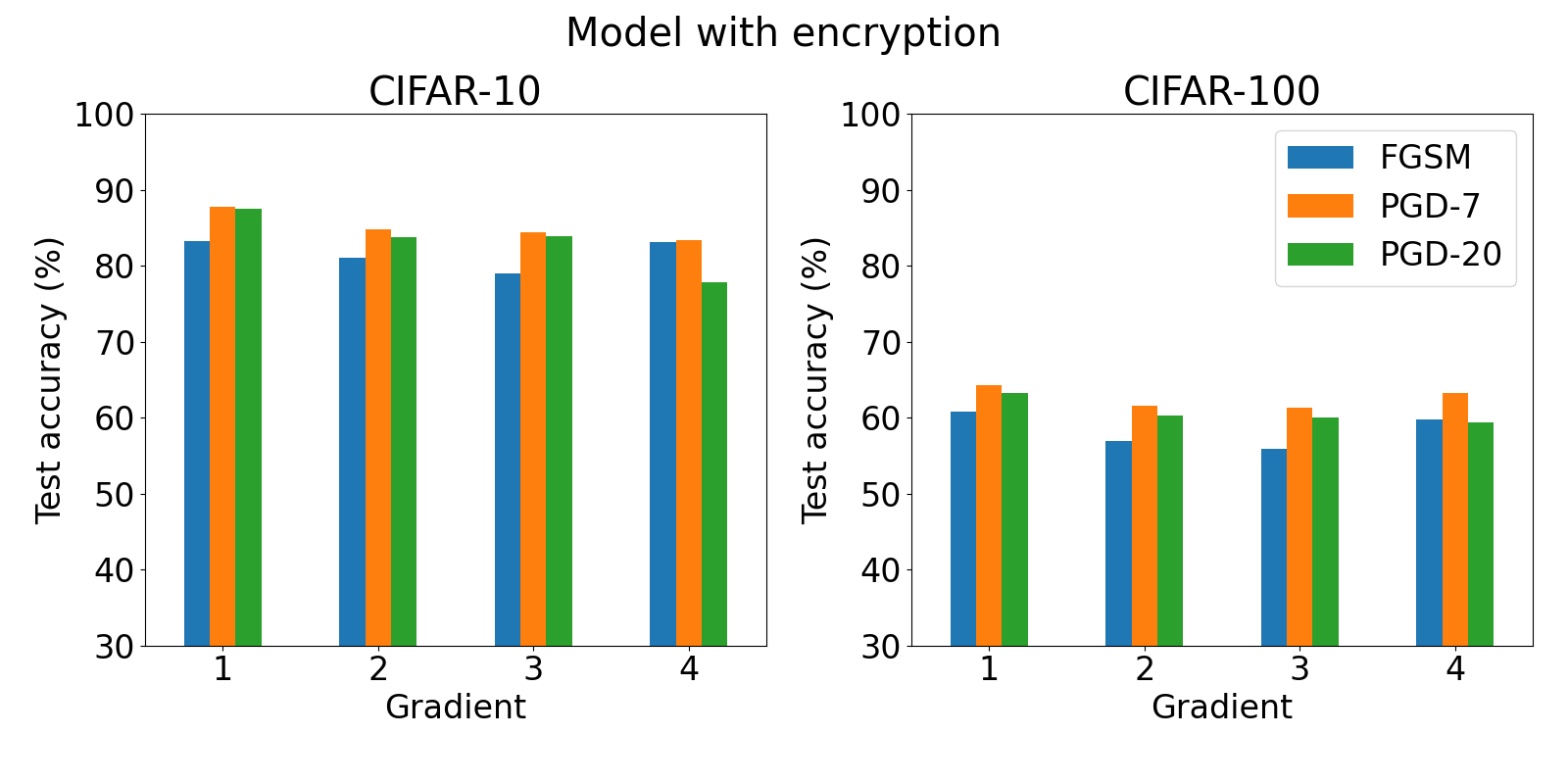}
    \caption{Our \textit{dark} gray box test accuracy against the adversarial images crafted using the 4 gradients. Top: the model without encryption is especially vulnerable to the PGD end-to-end gradient without encryption (gradient 2). Bottom: the model with encryption is resistant to all gradients.}
    \label{fig:cifar_pgd7_results_private_classifieranddenoiser}
\end{figure}

Table \ref{tab:graybox_all} shows our defense significantly outperforms other defenses in both natural and adversarial accuracy. Particularly, our natural accuracy is close to the one obtained by natural training. Further, Figure \ref{fig:cifar_pgd7_results_private_classifieranddenoiser} shows our individual adversarial accuracy against each gradient for each attack. In this figure, the top row shows that the model without encryption is especially vulnerable to the PGD end-to-end gradient without encryption (gradient 2). Here, the difference in the attacker's and defender's models is only their trained weights due to different weight initializations. In comparison, the bottom row shows that the model with encryption gains strong protection against all gradients. 

These results lead to the following conclusions. First, initializing the model weights using a private seed provides moderate protection against adversarial attacks. Second, using a denoiser to remove the potential adversarial noise drastically improves the natural and FGSM accuracy. However, without encryption, this defense is still vulnerable to the PGD attacks. Third, in comparison, using encryption slightly affects the natural and FGSM accuracy, but significantly boosts the PGD accuracy by 12 to 19\%.

\subsubsection{Key Length} \label{sec:keylength}

\begin{figure*}[h]
    \centering
    \includegraphics[width=\textwidth]{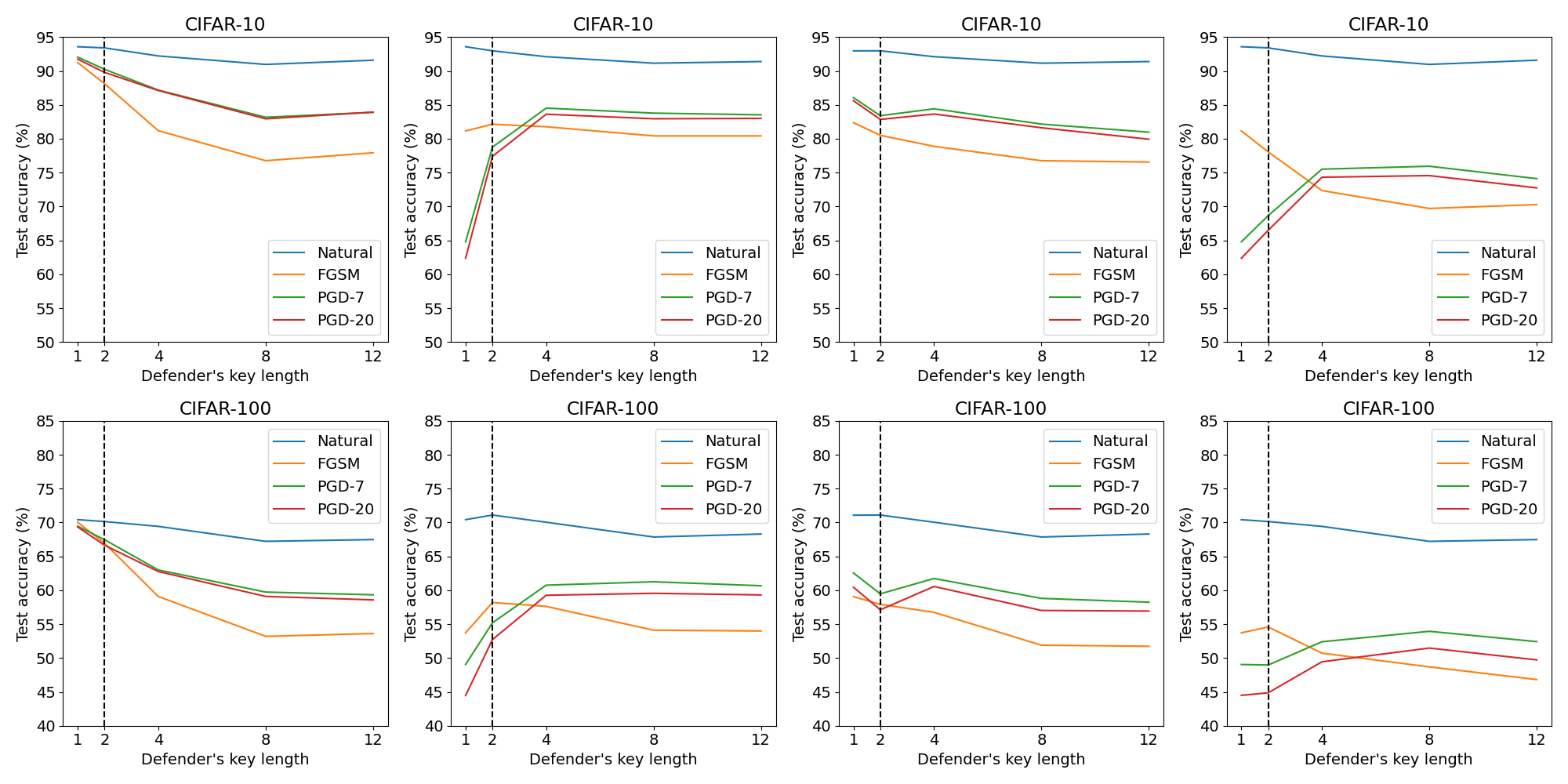}
    \caption{Our gray box test accuracy against the adversarial attacks crafted using different gradients, where the defender's key length $m=1,2,4,8,12$. The vertical dotted line separates $m=1$ from the rest, since it corresponds to the model without encryption. 1st column: the attacker uses the gradient from the classifier. 2nd column: the attacker uses the end-to-end gradient without encryption. 3rd and 4th columns: the attacker uses the end-to-end gradient with encryption; 3rd column: the attacker and defender use different keys, 4th column: they use the same key for each key length.}
    \label{fig:cifar_pgd7_keylength_allresults}
\end{figure*}

We discuss the impact of the defender's key length $m$ on the gray box test accuracy, where $m = 1,2,4,8,12$. These keys are shown in Table \ref{tab:keys_keylength}, and they are randomly chosen. For $m=$ 1 and 2, there is only one possible key. $m=1$ is a special case, since encrypting using the key $(32)$ is equivalent to the model without encryption. This means that the shortest key length for encryption is 2. 

For these experiments, the attacker and defender initialize their models with different random seeds. We use the first three gradients in Section \ref{sec:attacks} to craft the adversarial images (for gradient 3, the attacker's key is randomly chosen as $\texttt{k}' = (2, 16, 8, 4, 1, 1)$). To evaluate the benefit of the key, we include an additional case where the attacker and defender use the same key for each key length; this means their encrypted inputs are the same, and the difference is only their trained weights due to different weight initializations.

\begin{table}[h]
    \centering
    \begin{tabular}{|c|c|} 
     \hline
     Defender's key & Key length $m$ \\
     \hline
     $(32)^\star$ & $1$ \\ 
     $(16, 16)^\star$ & $2$ \\ 
     $(4, 4, 8, 16)$ & 4 \\ 
     $(1, 8, 4, 4, 2, 4, 8, 1)$ & 8 \\ 
     $(2, 1, 1, 1, 2, 8, 2, 4, 2, 1, 4, 4)$ & 12 \\ \hline
    \end{tabular}
    \caption{The defender's keys. $\star$ indicates it is the only key for $m$.}
    \label{tab:keys_keylength}
\end{table}

Figure \ref{fig:cifar_pgd7_keylength_allresults} shows that $m$ does not have a significant impact on our gray box test results with encryption. Further, if the defender's key is different from the attacker's (column 3), the adversarial accuracy is much better compared to the case when their keys are the same (column 4), which demonstrates the effectiveness of the key.

\subsubsection{Light and Medium Gray Box Defenses} \label{sec:pub_cls}

This section presents the test results when the defender's classifier $f$'s weights are public. For adversarial training, this is equivalent to the \textit{white} box defense, since the attacker can obtain the gradient directly from $f$. In contrast, our defense is \textit{light} or \textit{medium gray} box since the key is private, so the attacker cannot obtain the end-to-end gradient directly from the defender's model. Thus, it is not entirely fair to compare the results. Instead, our goal is to show that as more components in the defender's model become private, the adversarial accuracy considerably improves. To attack our defense, the 4 gradients in Section \ref{sec:attacks} are used to craft the adversarial images, and our minimum accuracy is reported.

\begin{figure*}[h]
    \centering
    \includegraphics[width=0.7\textwidth]{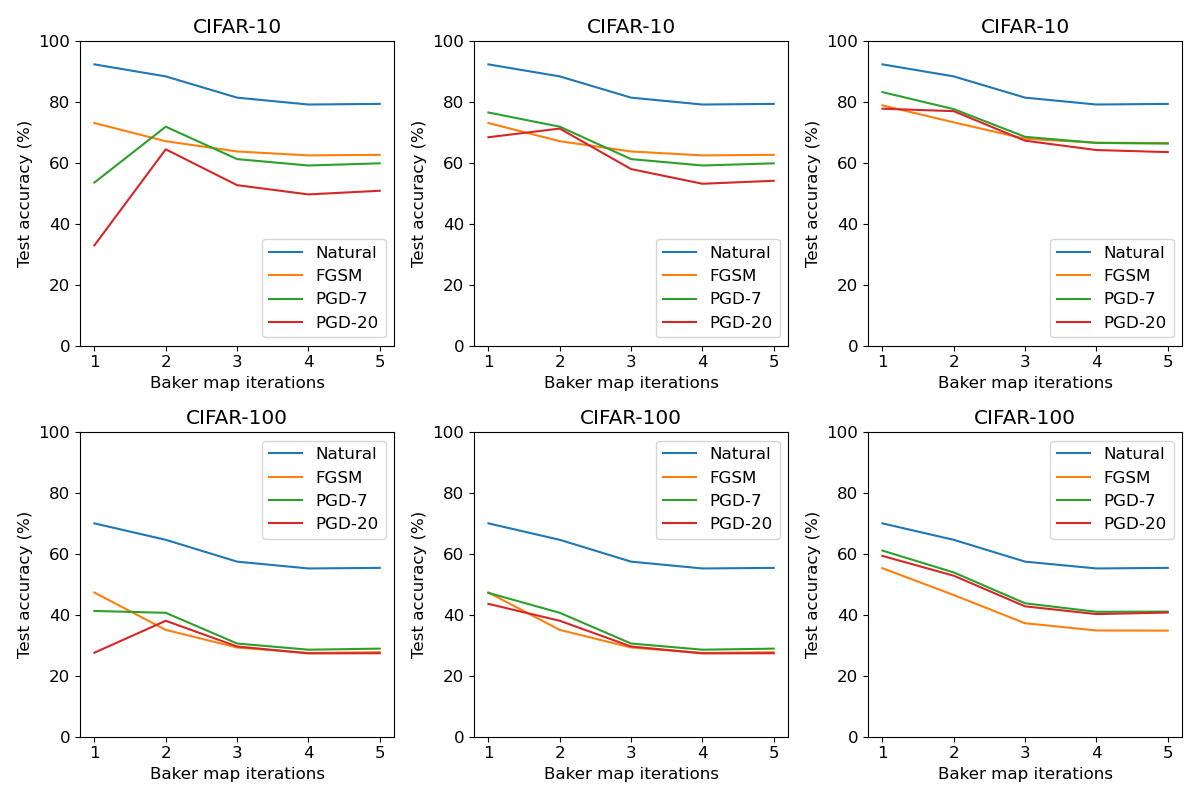}
    \caption{Test accuracy of our gray box defenses vs. the number of Baker map iterations. 1st column: \textit{light} gray box. 2nd column: \textit{medium} gray box, where the classifier's and denoiser's weights are public and private. 3rd column: \textit{dark} gray box. In the 1st case, the optimal number of Baker map iterations is 2 due to PGD attacks. In the 2nd and 3rd cases, the optimal number is 1.}
    \label{fig:cifar_test_min_accs}
\end{figure*}

\begin{figure}[h]
    \centering
    \includegraphics[width=\columnwidth]{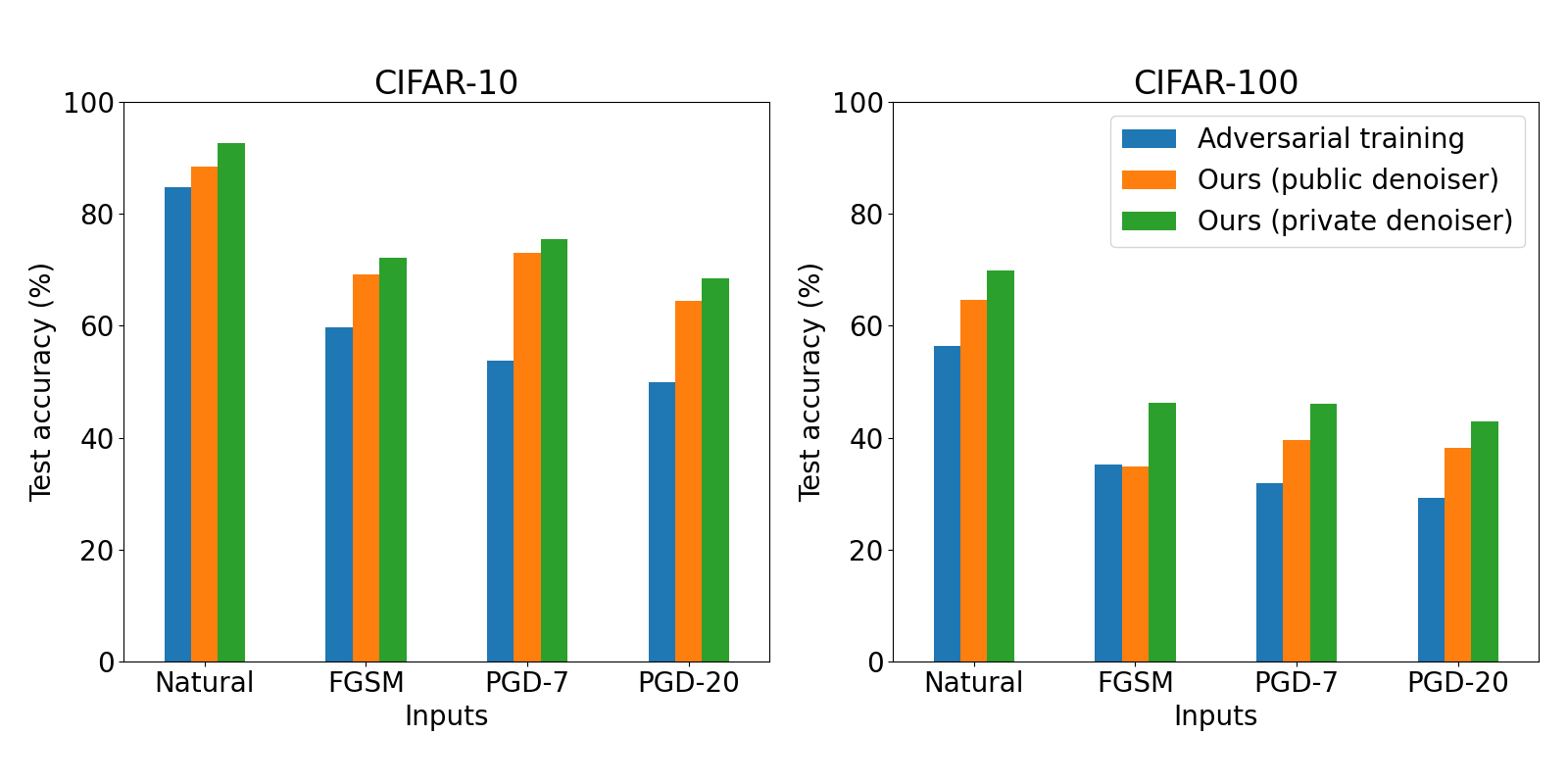}
    \caption{Test accuracy when the defender's classifier's weights are public. Adversarial training, ours with public and private denoiser correspond to the \textit{white}, \textit{light gray} and \textit{medium gray} box defenses. In general, the performance improves as more model components become private. For ours, the discrepancy in the natural accuracy using the public vs. private denoiser is due to the number of Baker map iterations taken in each case.}
    \label{fig:public_classifier}
\end{figure}

Figure \ref{fig:cifar_test_min_accs} illustrates the test accuracy of our gray box defenses vs. the number of Baker map iterations. In general, as the iteration increases, the encrypted images become more random, and our performance decreases; the PGD accuracy at iteration 1 is an exception, which we will explain in the following text. Figure \ref{fig:public_classifier} shows that for our defense, when both the classifier's and denoiser's weights are public (i.e., \textit{light} gray box), the performance is already much better than adversarial training, except for FGSM on CIFAR-100, which is slightly worse. Here, if we use only 1 Baker map iteration, the model is especially vulnerable to the PGD end-to-end gradient with encryption using random keys (gradient 4); namely, the attacker uses the defender's model weights, and samples a random key for each PGD step. However, if 2 iterations are used, the method can effectively defend against gradient 4, although it comes with a small penalty for the natural and FGSM accuracy. We conjecture that by using 2 iterations, the defender's encrypted inputs are sufficiently random, such that it is difficult for the attacker to sample similar inputs to craft the adversarial images. Meanwhile, when the denoiser's weights are private (i.e., \textit{medium} gray box), gradient 4 is no longer effective for the attack, so we use 1 Baker map iteration, and the results are better compared to the \textit{light} gray box case. In sum, the defender's performance improves as more model components become private.

\section{Related Work}

Our proposed method is a deterministic defense controlled by a random seed, and the seed determines the transformation of the inputs as well as the initialization of the model weights. Previous works such as \cite{buckman,guo} use input transformations as a defense, but their goal is to shatter the gradient, such that the attacker cannot find the correct gradient to craft adversarial images. There are also defenses based on test-time randomization, such as \cite{xie,dhillon}, which ensure that the defender's inputs and/or model weights are different from the attacker's during inference. Unfortunately, \citet{obfuscate} show that all these defenses can be broken by approximating the gradient for the defender's model. In comparison, for our defense, unless the attacker knows the defender's private seed, it is difficult to approximate the gradient due to the Baker map encryption and random weight initializations. Lastly, other interesting key-based gray box defenses include \cite{chen,vinh}.

\section{Conclusion and Future Work}

The previous literature on adversarial examples focused heavily on white box and black box attacks, but paid little attention to the equally practical gray box attacks. We propose a novel defense that assumes everything but a private key will be made available to the attacker. Our framework uses an image denoising procedure coupled with encryption via a discretized Baker map. Extensive testing against the FGSM and PGD adversarial images crafted using various gradients shows that we achieve significantly better results than the state-of-the-art gray box defenses in both natural and adversarial accuracy. Our method is easy to implement, suitable for high-resolution inputs, and efficient in testing.

To prevent the black box attacks where the attacker trains a surrogate model to mimic the behavior of defender's, the defender can train an ensemble of denoisers, each using a different private key (the classifier is the same). Then, each time in testing, they randomly choose one denoiser from the ensemble to make the prediction. Since the total number of encryption keys is very large, it is infeasible for the attacker to train an ensemble of denoisers using all keys, and then use the mean gradient for the attack. We leave the verification of this idea for future work.

\paragraph{Acknowledgements} This research was supported by the NWO Perspective grant DLMedIA and the in-cash and in-kind contributions by Philips.

\bibliography{mybibliography}
\bibliographystyle{icml2021}


%%%%%%%%%%%%%%%%%%%%%%%%%%%%%%%%%%%%%%%%%%%%%%%%%%%%%%%%%%%%%%%%%%%%%%%%%%%%%%%
%%%%%%%%%%%%%%%%%%%%%%%%%%%%%%%%%%%%%%%%%%%%%%%%%%%%%%%%%%%%%%%%%%%%%%%%%%%%%%%

\end{document}